# ANN-based Innovative Segmentation Method for Handwritten text in Assamese

## Kaustubh Bhattacharyya and Kandarpa Kumar Sarma

Department of Electronics and communication technology, Gauhati University,
Guwahati, Assam ZIP: 781014, India

## Abstract

Artificial Neural Network (ANN) s has widely been used for recognition of optically scanned character, which partially emulates human thinking in the domain of the Artificial Intelligence. But prior to recognition, it is necessary to segment the character from the text to sentences, words etc. Segmentation of words into individual letters has been one of the major problems in handwriting recognition. Despite several successful works all over the work, development of such tools in specific languages is still an ongoing process especially in the Indian context. This work explores the application of ANN as an aid to segmentation of handwritten characters in Assamese- an important language in the North Eastern part of India. The work explores the performance difference obtained in applying an ANN-based dynamic segmentation algorithm compared to projection- based static segmentation. The algorithm involves, first training of an ANN with individual handwritten characters recorded from different individuals. Handwritten sentences are separated out from text using a static segmentation method. From the segmented line, individual characters are separated out by first over segmenting the entire line. Each of the segments thus obtained, next, is fed to the trained ANN. The point of segmentation at which the ANN recognizes a segment or a combination of several segments to be similar to a handwritten character, a segmentation boundary for the character is assumed to exist and segmentation performed. The segmented character is next compared to the best available match and the segmentation boundary confirmed.

***Keywords:*** *Segmentation, Classification, Handwritten, Cursive, Recognition, Dissection.*

## 1. Introduction

Artificial Neural Network (ANN)s have been preferred tools for pattern recognition including optical characters which broadly constitutes an important segment of Computer Vision and Machine Learning. This is because ANNs have the capacity to learn, adapt to changing environments and demonstrate a computational paradigm that resembles the parallelism generated by the human brain. That way ANNs are smart tools and can be applied for a host of pattern recognition and prediction problems [1]. ANNs have two phases of working- first training during which it learns the patterns and testing which ascertains the extent of learning.

Optical Character Recognition (OCR) is a common and popular example of application of ANNs for pattern recognition. OCR refers to a process of generating a character input by optical means, like scanning, for recognition in subsequent stages by a smart tool like ANN by which a printed or handwritten text can be converted to a form which a computer can understand and manipulate. A generic character recognition system may be shown in figure 1. Its different stages are as below:

- **Input:** Samples are read to the system through a scanner.
- **Preprocessing:** Preprocessing converts the image into a form suitable for subsequent processing and feature extraction.
- **Segmentation:** The most basic step in OCR is to segment the input image into individual *glyphs*. This step separates out sentences from text and subsequently words and letters from sentences.
- **Feature extraction:** Extraction of features of a character forms a vital part of the recognition process. Feature extraction captures the vital details of a character.
- **Classification:** During classification, a character is placed in the appropriate class to which it belongs. Character classification is roughly categorized as *Sub-symbolic classifiers* and *Symbolic classifiers*. The ANN approach is classified to be a sub symbolic classifier [2].

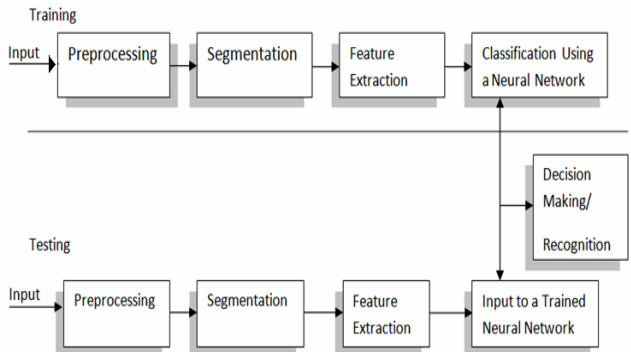

Fig-1 Block diagram Character Recognition System

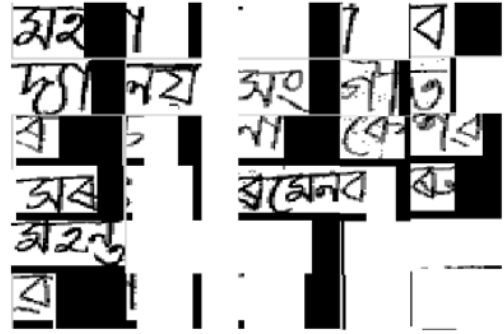

Figure 2: Segmented output of the hand written characters

As mentioned in [3], [4], segmentation plays an important role in the overall process of recognition of printed and handwritten characters. This is more so with **cursive** writing. Success and failure of an OCR system depends on the segmentation process. But this description is related to a work that attempts to use ANNs for the segmentation stage of an ANN based OCR system exclusively for Assamese which is an important language in NE region of India. The reasons behind the use of ANNs for segmentation are as below:

1. Static segmentation method suffers from a serious drawback that it cannot fix segmentation boundaries for cases where inputs have size and inclination variations.

2. Static Segmentation methods also fail to fix segmentation boundaries for cases where there are writer induced variations in inputs. Figure 2 shows the failure of static segmentation methods in dealing with writer induced variations.

The solution for such cases can be given by ANNs these have the ability to learn shapes and that way discriminate segmentation boundaries.

ANNs have been used for several character recognition systems. Some of the segmentation methods relevant in practice is described in [3]. For cursive writing Cheng, Liu et. al [5] provides a description of available segmentation methods. Use of ANNs for segmentation has been reported by Blumenstein [6]. Other similar works are [7], [8], [9], [10], [11], [12], [13] to name a few. Very few known attempts have been reported regarding use of ANNs for segmentation in the Indian OCR scenario.

Section 2 provides an insight into certain features of Assamese scripts. Details of experimental work have been included in section 3. Use of ANNs for segmentation and related description has been covered in section 4. Section 5 concludes the description.

## 2 Basic features of Assamese Handwritten Characters:

1. Assamese characters can trace their roots to Brahmi script and has evolved over the years through modifications.

2. Assamese script formed by 11 vowels, 40 consonants, over 10 modifiers and over 300 compound characters. The use of upper and lower case letters like in English is not there in Assamese as in other languages including Bengali.

3. There is a use of head line (called matra in Assamese) in certain characters including consonant and vowels. It helps in segmentation of the characters easily but as many of the segmented characters with their head lines missing appears similar making classification difficult.

4. A typical Assamese word maybe classified into three Jones as in Figure 3

- **Upper Zone** Area above the head line. It is characterized by the presence of extensions of the modifiers.

- **Middle Zone** Area where the main body of the character lies.

- **Lower Zone** Area where some of the modifiers exists.

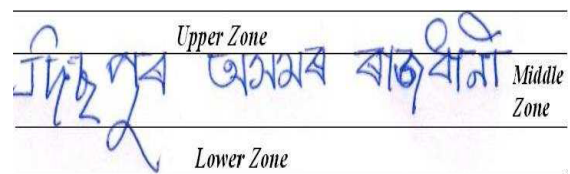

Figure 3: Assamese words with three different zones

## 3. Detail of the Experimental Work

The work conceptualized an ANN based OCR system where segmentation is done by a multi-layered perceptron (MLP)- a class of feed forward neural network. The MLP is trained to do so. The algorithm involves, first training of an ANN with individual handwritten characters extracted from different individuals. Handwritten sentences are separated out from text using a static segmentation method. From the segmented line, individual characters are separated out by first over segmenting the entire line. Prior to all these steps some preprocessing steps are required for the scanned image. These are:

**1. Noise removal:** It involves noise removal using certain filtering operations.

**2. Enhancement:** Here the filtered images are enhanced using certain high boost filter makes and histogram equalization technique.

**3. Sharpening:** For degraded or blurred images after noise cleaning operations sharpening may be done.

**4. Binarisation:** After enhancement and sharpening the gray level image is converted into binary form so as to ease the computational load of the subsequent stages

**5. Normalization:** The images just before the segmentation stage are converted to certain standard sizes. If the input has inclination and skew, respective corrections are done. After preprocessing the next step is segmentation of the input. During this stage first lines are separated out from the text first into lines and then the words are next segmented into the individual characters. The approach adopted here is a projection -based one. A brief outline of the static segmentation method is as below:

**1. Row -wise dissection:**
- *C*alculate row-wise pixel sums of the inputs
- *O*btain the row -wise projection of the inverted inputs
- *F*ind the minimum of the projections
- *A* pair of closely lying minimum points defines one segmentation boundary.
- *D*issection boundaries give sub-images of inputs. Hold them in an array.

**2. Column-wise dissection:**
For each entry into the array as above
- *C*alculate the sum of pixels column-wise.
- *O*btain the column-wise projection of the inverted sub-images.
- *F*ind the minimum points from these projections
- *A* pair of consecutive minimum points defines one segmentation boundary.
- *S*tore every character dissected out of the sub-image into an array. The array must also include space in between words.
- *T*he array holds the segmented outputs of the segmented sub-images as obtained in step 1.

The results obtained are shown in Figures 2, 7 and 13. In certain cases character spacing is non-uniform, after the head-lines are removed. Base line character spacing then becomes comparable to word spacing. This affects word spacing. In such a situation morphological dilation maybe used as described in [14]. In case, modifiers are not separated from characters, especially in the case where modifiers are lying below the middle zone i.e. in the lower zone, the statistics of the horizontal projection is so obtained that a threshold is fixed that is 1.5 times the average line height. The non-zero valleys below the threshold indicate the separation boundary between the character and the modifier [14].

This method has certain drawbacks which are described in subsequent stages.

### 3.1 Similarity Measure

A similarity measure for the machine printed characters may be defined as:

$$S = \left(1 - \frac{\sum I_{seg}(i,j)}{\sum I_{ref}(i,j)}\right) * 100 \qquad (1)$$

Where $I_{seg}(I,j)$ is the segmented image and $I_{ref}(i;j)$ is the reference image. For touching characters the segmentation method suffers and the similarity measures show lesser values. The segmentation method is not suitable for touching characters (Figure 2) and is useful more for printed characters which are a bit isolated (Figure 7). The method is useful in separating modifiers in the case where characters as well (Figure 13). This is shown by the final segmented result of italic characters. The segmentation method is however capable of segmenting modifiers from the consonants despite their presence below and above the middle zone. Modifiers are decremented out to isolate the characters before feature extraction. The headline, similarly, can be segmented out which makes separation of the characters from individual words simpler. But this process also has the possibility of completely losing certain vital information regarding the characters.

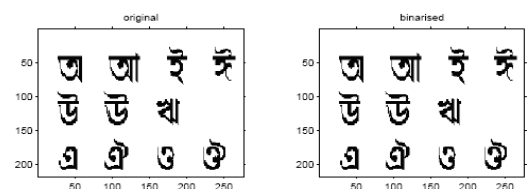

Figure 4: Pre-processed input of vowels

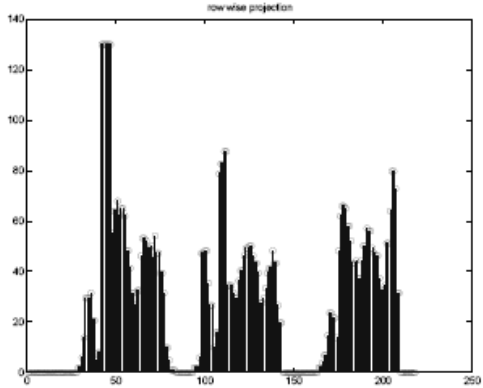

Figure 5: Horizontal Projection of the vowels

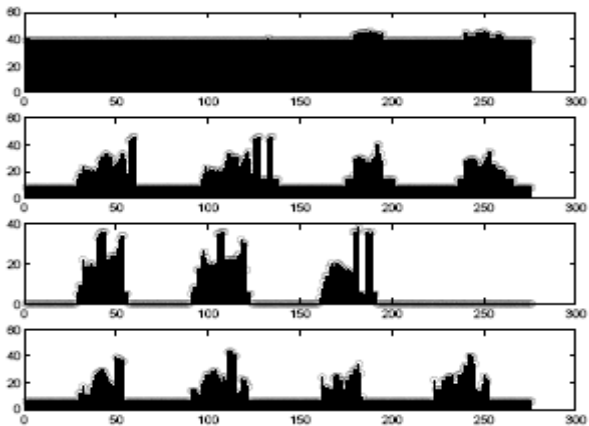

Figure 6: Vertical segmentation of vowels

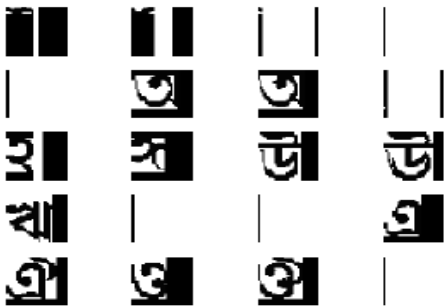

Figure 7: Segmented output of the vowels

The values of similarity measure for different segmented characters are shown as in Table-1. A higher value of the similarity measure directly improves the classification and recognition rates subsequently. An improved method for segmentation may yield higher values of the similarity measure than has been found in the present work. The result obtained from static segmentation was found not to be healthy for handwritten characters (Figure 2). The basis of the ANN based segmentation method has been developed keeping into account the failure shown by the static method (Figure- 2). The steps of ANN based segmentation may be described as below.

1. ANN trained with all available Assamese characters.

2. Sentences are separated out from text using static method.

- On sentences thus obtained an over segmentation is done. The over segmentation is done at an interval of 2:5% length to that of that total length of the sentence separated (fig 8).
- Each of the segment obtained from the sentence due to the over segmentation (fig 8) are fed to the trained ANN.
- If due to the feeding of first segment the ANN fails to recognize, the next segment is fed together. This process is repeated till the ANN recognizes a complete character (Figure 9).
- The recognition performances of the trained ANN then are determined. A successful recognition by the ANN provides a demarcation and generates a segmentation boundary.
- This process is repeated till the segmentation boundaries for the entire sentence and then the text is completed.

Table 1: Similarity Measure values obtained using static segmentation

| Pronunciation | S in % |
|---|---|
| A | 86 |
| AA | 82 |
| E | 92 |
| EE | 92 |
| U | 92 |
| UU | 92 |
| RI | 82 |
| AE | 90 |
| AOI | 90 |

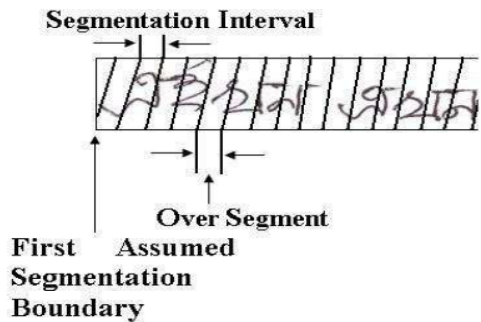

Figure 8: Segmentation boundary Measure

The entire work maybe depicted by the figure 14.

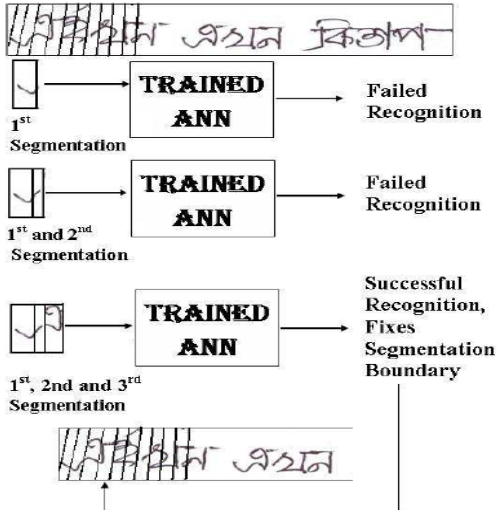

Figure 9: Segmentation Process Using trained ANN

# 4. ANN Based Segmentation

First the ANN is trained with individual **Assamese** characters. This ANN will handle the recognition part of the segmentation process. To use an ANN for segmentation, it must be first trained. The training of the ANN takes into account the configuration of the ANN.

## 4.1 Configuration of MLP's

Several configurations of the MLP were utilized for training. These had one and two hidden layer configurations over and above the input and output layers. The one hidden layered configuration emerged as a trade-off between computational complexity and performance.
The three layered MLP i.e, the one with one hidden layers had the following configuration:
Length of input layer = Normalized size of the input character,
Length of first hidden layer = 1.5 times of the feature vector,
Length of output layer=number of classes;
The two, three and four hidden layer MLP similarly had the varying hidden layer length configuration. The choice of the length of the hidden layers has not followed any definite logic but has been randomly selected.
Several methods of (Error) Back Propagation have been used for training the ANN. These are:

- Gradient Descent (GDBP).
- Gradient Descent with Momentum (GDMBP)
- Gradient Descent with Adaptive Learning rate Back propagation (GDALBP)
- Gradient Descent with Momentum and Adaptive Learning Rate BP(GDMALRBP)

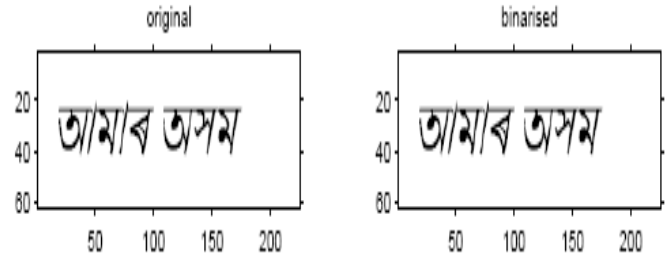

Figure 10: Pre-processed input of italic characters

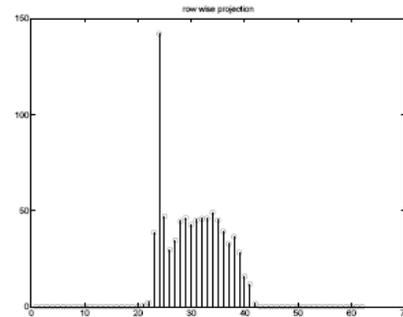

Figure 11: Horizontal Projection of the italic characters

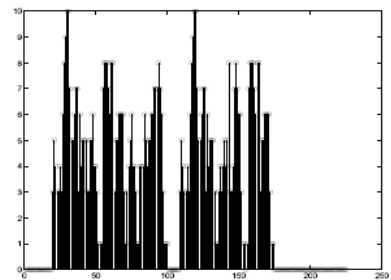

Figure 12: Column-wise segmentation of the italic characters

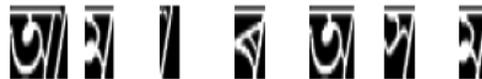

Figure 13: Segmented output of the italic characters

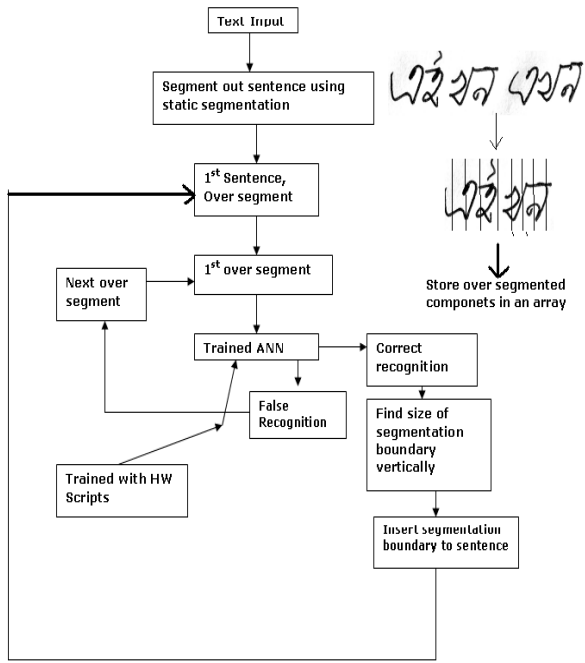

Figure 14: Detail Steps of the Present Work

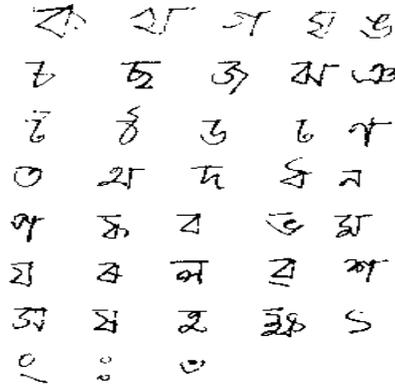

Figure 15: A section of characters forming training set of ANNs used for segmentation

The training input to the ANNs used for segmentation is a set of handwritten scripts a part of which is shown in Figure 15. The complete training set includes variations of twenty five different persons. For a three layered MLP the mean square error (MSE) attained after 1000 to 4000 training sessions is    depicted in Table 2. Table 2 shows that the 3-layered MLP trained with GDMALRBP attains the best MSE convergence. Table 3 shows the classification performance of a three layered MLP trained with the four mentioned training methods. The GDMALRBP based training to the ANN generates the best classification performance. The ANN trained by following these considerations is taken for performing the segmentation of input handwritten scripts. The number of training sessions has been restricted between 1000 to 4000 training sessions. This is because with training epochs below 1000 the MLPs don't develop the ability to make discrimination between classes with success rates above 50%. Again with training sessions over 4000, there is always a chance of the ANNs losing the ability to generalize classes.

The testing set includes handwritten scripts of twenty different persons writing in five different days. That way the testing set consists of over hundred samples. Moreover, noise has been added to check the versatility of the system and its ability to tackle variations in input conditions. A set of test samples generated by sentence-wise over-segmentation of the recorded samples using the static method and used to verify the dynamic segmentation ability of the selected MLP is shown in Figure 16. The average results of the segmentation thus carried out is expressed by the similarity measure referred to earlier in section 3.1. A comparison of the similarity measures obtained by the static and ANN based segmentation methods are shown in Table 4. The advantage of the ANN-based method is obvious.

Table 2: MSE attained during training by a three layered MLP with a learning rate of 0:4

| Sessions | MSE attained with four different training methods | | | |
|---|---|---|---|---|
| | GDBP | GDMBP | GDALBP | GDMALRBP |
| 1000 | $10.1 \times 10^{-3}$ | $9.1 \times 10^{-3}$ | $7.6 \times 10^{-3}$ | $6.23 \times 10^{-3}$ |
| 2000 | $1.02 \times 10^{-3}$ | $0.81 \times 10^{-3}$ | $0.71 \times 10^{-3}$ | $0.52 \times 10^{-3}$ |
| 3000 | $0.2 \times 10^{-3}$ | $0.08 \times 10^{-3}$ | $0.06 \times 10^{-3}$ | $0.04 \times 10^{-3}$ |
| 4000 | $0,05 \times 10^{-3}$ | $0.02 \times 10^{-3}$ | $0.01 \times 10^{-3}$ | $0.008 \times 10^{-3}$ |

Table 3: Classification performance in % by a three layered MLP trained with GDMALRBP with a learning rate of 0:4

| Sessions | Classification rate in % four different training methods | | | |
|---|---|---|---|---|
| | GDBP | GDMBP | GDALBP | GDMALRBP |
| 1000 | 87.1 | 90.3 | 91.3 | 92.2 |
| 2000 | 86.9 | 91.2 | 92.2 | 93.7 |
| 3000 | 89.1 | 92.3 | 93.6 | 95.4 |
| 4000 | 89.9 | 93.4 | 94.6 | 95.5 |

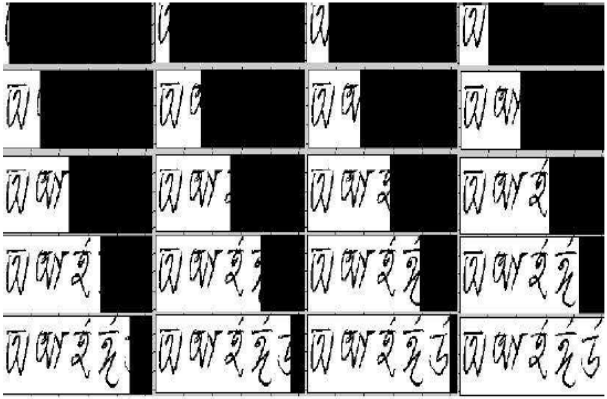

Figure 16: Testing set of ANNs used as combinations of over-segments

Table 4: Comparison of Similarity Measure (S) in % of five characters generated by static segmentation and by the selected ANN after 1000 to 4000 training sessions

| Cases | Pronunciations | S with Static Segmentation | S in % between 1000 to 4000 training sessions | | | |
|---|---|---|---|---|---|---|
| | | | 1000 | 2000 | 3000 | 4000 |
| 1 | A | 86 | 88.1 | 94.0 | 95.3 | 96.3 |
| 2 | AA | 82.0 | 84.4 | 90.1 | 92.2 | 93.2 |
| 3 | E | 92.0 | 92.3 | 92.9 | 94.0 | 95.0 |
| 4 | EE | 92.0 | 93.2 | 94.8 | 95.0 | 96.1 |
| 5 | U | 92.0 | 93.1 | 95.6 | 95.9 | 97.0 |

## 5. Discussion and Conclusion

The work offers an insight into development of a segmentation system for Assamese scripts using ANN. The work concentrates on improving the performance of the segmentation stage handling handwritten inputs with an aim to aid a character recognition system. The system tackles well non touching handwritten characters and even separates out partially touching characters without inclination but fails to tackle writing styles where characters are inclined and touching each other. Further development of the system can be a stage where touching handwritten characters can be handled by the system. The benefit of such a system is that it improves performance of segmentation which is so important in an OCR system. The ANN trained with different character shapes can be used as a segmentation making tool while selecting segmentation boundaries. As the segmentation boundary is fixed dynamically, the system can deal effectively with segmentation problems that suffer due to written induced variations in size and shape of characters without inclinations.

## About the Authors

**Kaustubh Bhattacharyya** is currently working as a Project Fellow under ANN based project in the department of Electronics and Communication Technology, Gauhati University, Assam. He obtained M.Phil in Electronics Science from Gauhati University, M.Sc in Electronic Science with DSP as a specialization from Gauhati University in 2007.

**Kandarpa Kumar Sarma** is currently working as a lecturer, Department of Electronics and Communication Technoly, Gauhati University, Assam. He is working towards PhD in Biomedical Signal Processing at ECE, IIT Guwahati. He obtained M.Tech in Digital Signal Processing from IIT Guwahati with specialization in Pattern Recognition and ANN, M.Sc (Electronics) from Gauhati University in 1997. He has published 21 research papers in reputed conferences, and carried out research project worth Rs. 32 lakhs so far. He is a member of Bio Sensor Society of India, Instrumentation Society of India and Indian Physical Academy.